\documentclass[letterpaper, 10 pt, conference]{ieeeconf}  

\usepackage{acro}

\usepackage{algorithm} 
\usepackage{algorithmic} 

\usepackage{array} 
\usepackage{booktabs} 
\usepackage{blkarray}
\usepackage{multirow} 

\usepackage{subcaption} 

\usepackage{color} 
\usepackage[dvipsnames]{xcolor} 

\usepackage[utf8]{inputenc} 

\usepackage{epsfig} 
\usepackage{float} 
\usepackage{graphicx} 

\usepackage{setspace} 
\usepackage{stfloats} 

\usepackage{amsmath} 
\usepackage{amssymb} 
\usepackage{amsfonts} 

\usepackage{cite} 
\usepackage{hyperref} 

\usepackage{textcomp} 
\usepackage{lipsum} 
\usepackage{soul} 

\usepackage{ifpdf} 
\usepackage{url} 
\usepackage{verbatim} 
\usepackage{optidef}
\usepackage{bm} 

\usepackage[table]{xcolor}
\definecolor{mygray}{rgb}{0.9,0.9,0.9} 
\usepackage{threeparttable}
\usepackage{nicematrix}

\usepackage{balance}
\usepackage{graphicx}  
\usepackage{subcaption} 
\usepackage{caption}

\usepackage{amsthm}

\let\labelindent\relax
\usepackage{enumitem}

\usepackage{fancyhdr}

\definecolor{lightgray50}{gray}{0.93} 

\newcommand{\pr}[1]{\ref{prop:#1}}
\DeclareAcronym{ACT}{short = ACT,
	long = Action Chunking Transformer
}

\DeclareAcronym{RTFF}{short = RTFF,
	long = Random-to-Target Fabric
Flattening
}

\DeclareAcronym{VS}{short = VS,
	long = Visual Servoing
}

\DeclareAcronym{IL}{short = IL,
	long = Imitation Learning
}

\DeclareAcronym{CAD}{short = CAD,
	long = Computer Aided Design
}

\DeclareAcronym{GAT}{short = GAT,
	long = Graph Attenuation Network
}

\DeclareAcronym{DoF}{short = DoF,
	long = Degrees of Freedom
}

\DeclareAcronym{EE}{short = EE,
	long = End-Effector
}

\DeclareAcronym{MACT}{short=MACT,
    long = Mesh Action Chunking Transformer
}

\DeclareAcronym{GT}{short=GT,
    long = Ground Truth
}

\theoremstyle{definition}
\newtheorem{definition}{Definition}

\IEEEoverridecommandlockouts                              
                                                          
\title{\LARGE \bf
RTFF: Random-to-Target Fabric Flattening Policy \\using Dual-Arm Manipulator
}

\author{Kai Tang$^{1,\dagger}$,
  Dipankar Bhattacharya$^{2,\dagger}$, Hang Xu$^{3}$,
  Fuyuki Tokuda$^{4}$, \\Norman C. Tien$^{1}$,
  and Kazuhiro Kosuge$^{5}$
\thanks{$^{\dagger}$These authors contributed equally to this work.}%
\thanks{$^{1}$Kai Tang and Norman C. Tien are with the Department of Electrical and Electronic Engineering, Faculty of Engineering, The University of Hong Kong, Hong Kong SAR, China. (E-mail: k.tang98@outlook.com)}
\thanks{$^{2}$Dipankar Bhattacharya is with the Dyson School of Design Engineering, Imperial College London, London, United Kingdom.}
\thanks{$^{3}$Hang Xu is with JD.COM, Shenzhen, China.}
\thanks{$^{4}$Fuyuki Tokuda is with the Unprecedented-scale Data Analytics Center, Tohoku University, Sendai 980-0845, Japan, and also with the Graduate School of Information Sciences, Tohoku University, Sendai 980-0845, Japan.}
\thanks{$^{5}$Kazuhiro Kosuge is with the Department of Mechanical Engineering, City University of Hong Kong, Hong Kong SAR, China.}
}
\begin{document}

\maketitle
\thispagestyle{empty}
\pagestyle{empty}

\thispagestyle{fancy}
\fancyhf{}
\renewcommand{\headrulewidth}{0pt}
\fancyfoot[L]{\scriptsize\textbf{This work has been submitted to the IEEE for possible publication. Copyright may be transferred without notice, after which this version may no longer be accessible.}}

\begin{abstract}

    Robotic fabric manipulation remains challenging due to fabric deformability and occlusions from wrinkles and the manipulator. This paper defines Random-to-Target Fabric Flattening (RTFF) as the task of bringing a randomly wrinkled fabric to an arbitrary user-specified wrinkle-free target pose. RTFF requires simultaneous flattening and pose alignment, where the two objectives are inherently coupled since flattening the fabric displaces its pose, while realigning it tends to introduce wrinkles. To solve this task, this paper anchors both the current and target fabric states to the same template mesh, enabling direct vertex-level wrinkle and pose assessment without registration. Building on this representation, a hybrid Imitation Learning--Visual Servoing (IL--VS) RTFF policy is proposed. A novel Mesh Action Chunking Transformer (MACT) leverages structured mesh observations to achieve goal-conditioned coarse alignment from a compact demonstration set, after which VS ensures precise convergence to the target. The policy is validated on a real dual-arm teleoperation system, demonstrating precise alignment to unseen target poses, fabric types, and scales. Code and videos: \url{https://kaitang98.github.io/RTFF_Policy/}

\end{abstract}
\section{Introduction}

\begin{figure}[t!]
	\centering
	\includegraphics[width=0.9\linewidth] {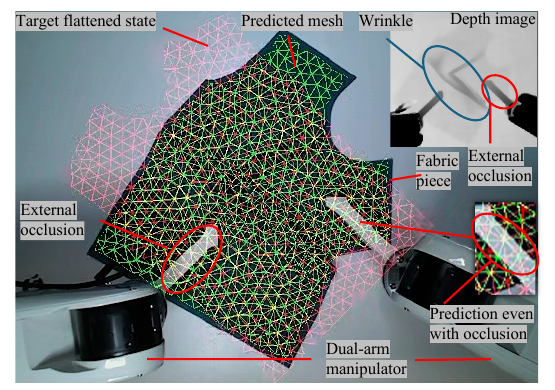} 
	\caption{\acf{RTFF} task.}
	\label{fig:introduction_figure} 
	\vspace{-1.7em} 
\end{figure}

Robotic manipulation of fabrics is a promising approach to reduce labor and enhance efficiency in garment production.
A key task is sewing, where the same fabric panel must be reloaded in different positions to stitch seams, and random wrinkles are inherently introduced during sewing, loading, and repositioning~\cite{Kai2025}.
To ensure accurate seam placement, fabric needs to be manipulated from an \textit{initial random wrinkled state} to a \textit{wrinkle‑free, flattened target state}, whose pose (orientation and translation) varies across processes.
This task, referred to as \textit{\ac{RTFF}} (Fig.~\ref{fig:introduction_figure}), requires iteratively selecting grasping points, simultaneously flattening local wrinkles, adjusting pose, and aligning the fabric until the target configuration is reached~\cite{taylor1997automated}.
Efficient RTFF is also important in fabric cutting, screen printing, and ironing, where flatness and precise alignment determine garment quality~\cite{GRIES2018179, Kai2025}.

Concretely, RTFF is a simultaneous flattening and pose-aligning problem, where the two objectives are \emph{inherently coupled} since flattening the fabric displaces its pose, while realigning it tends to introduce wrinkles. It must satisfy the following three properties: 
\begin{enumerate}[leftmargin=*, label=\textbf{(P\arabic*)}, ref=P\arabic*]
    \item \label{prop:1} \textbf{Arbitrary target flat pose.}
    The policy must accept a \emph{user-specified}, arbitrary flat pose (position and orientation), rather than a fixed canonical pose.

    \item \label{prop:2} \textbf{Simultaneous flattening and pose alignment.}
    The policy must \emph{jointly} remove all wrinkles across the entire fabric panel while aligning the panel to the target pose in a single unified framework.

    \item \label{prop:3} \textbf{Wrinkle-aware state estimation.}
    The policy must have real-time access to \emph{local} wrinkle states for flattening and the \emph{global} fabric pose for alignment.
\end{enumerate}

While prior work has made progress on related problems, such as fabric state estimation, flattening, and pose alignment, no existing method simultaneously satisfies all three requirements to achieve RTFF. 
Flattening and folding methods~\cite{Seita2020,hietala2022learning,tokuda2023cnn} ignore target pose~(\pr{1}) or provide only global feedback~(\pr{3});
alignment methods~\cite{tokuda2025transformer,shetab2023lattice,Kai2025}
assume a wrinkle-free start with given manipulation points~(\pr{2}); and 3D state estimation
methods~\cite{tang2022track,lin2021learning,huang2023self,longhini2025cloth} fit each observation independently and cannot compare directly to a target pose without a registration step~(\pr{1}, \pr{2}, \pr{3}). A detailed review of related work is provided in Sec.~\ref{sec:related}. In summary, the major limitation of existing methods is the lack of shared geometric reference between the observed fabric and the target pose.

Hence, this work proposes to define both the current fabric state and the target state with the same \textit{template mesh}, derived from the fabric's CAD model that deforms to match any observed state while preserving vertex identity. Prior work~\cite{wang2024trtm} applied template meshes for state estimation.
Building on this formulation, this paper proposes a template mesh-based hybrid IL--VS RTFF policy~(\pr{2}): (i) an \ac{IL} stage that learns from a small demonstration set to perform coarse alignment and wrinkle reduction guided by the mesh state, (ii) followed by a \ac{VS} stage that activates once the mesh signals sufficient flatness and drives the fabric to precise alignment with the target.
The contribution of this paper is summarized as follows:

\begin{itemize}

    \item A template mesh-based RTFF formulation that anchors both the current and target fabric states to the same canonical mesh, enabling direct vertex-level wrinkle and pose assessment~(\pr{1}, \pr{3}), and a hybrid IL--VS framework that enables simultaneous fabric flattening and pose alignment~(\pr{2}).

    \item A goal-conditioned \ac{IL} policy, \ac{MACT}, that leverages the template mesh state representation to achieve RTFF from a compact demonstration set.
    
    \item Physical validation on a dual‑arm teleoperation system, demonstrating precise alignment to unseen target configurations and robustness across different fabric types and scales.

\end{itemize}

\begin{figure*}[h!]
    \centering
    \includegraphics[width=\linewidth]{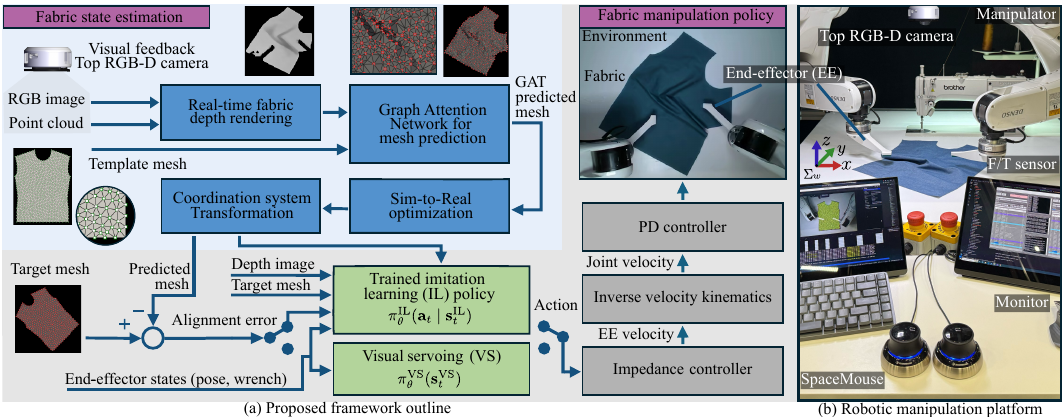}
    
    \caption{Overview of the proposed RTFF policy framework.}
    \label{fig:overall_framework}
    \vspace{-1.7em} 
\end{figure*}

\section{Related Work}
\label{sec:related}

\subsection{Fabric flattening and folding}
Pick-and-place policies trained from visual observations learn to reduce wrinkles by maximizing a global flatness metric such as image coverage~\cite{Seita2020}. Dynamic folding controllers~\cite{hietala2022learning, lin2021learning} use reinforcement learning with a global shape reward to execute pre-defined fold targets. While effective at these tasks, these methods do not accept a user-specified arbitrary target pose~(\pr{1}) and cannot ensure locally wrinkle-free~(\pr{3}), since their objectives use global/sparse rewards rather than local wrinkle state.

\subsection{Fabric alignment and \ac{VS}}
\ac{VS} methods reposition fabric toward a goal configuration using Jacobian-based control with a rigidity-preserving lattice or mesh model~\cite{shetab2023lattice, Edmund2025} or
Transformer-predicted texture-match error~\cite{tokuda2025transformer}.
Both require the fabric to be approximately flat or held mechanically taut. However, wrinkles in RTFF break the near-rigid deformation model and the gripper-tension assumption on which these methods rely~(\pr{2}). CNN-based servoing~\cite{tokuda2023cnn} does address simultaneous positioning and flattening~(\pr{2}), but it converges to a single fixed target baked into the CNN weights, hence it cannot accept an arbitrary user-specified pose at inference time~(\pr{1}), and provides no per-region wrinkle feedback~(\pr{3}).

\subsection{Fabric state estimation}
Fabric state has been represented using pixels~\cite{tokuda2023cnn,tokuda2025transformer},
edge features~\cite{tang2024time}, and latent vectors~\cite{yan2020learning}, but these 2D representations provide no 3D geometric structure and cannot assess
wrinkle severity or pose offset relative to a target.
Richer geometric representations improve reconstruction fidelity:
point-cloud tracking~\cite{tang2022track} and per-step mesh
fitting~\cite{lin2021learning,huang2023self} reconstruct cloth deformation from
depth or RGB-D inputs, while Gaussian splatting~\cite{longhini2025cloth} further
improves photometric accuracy.
However, these methods estimate each state independently, without
persistent vertex correspondence shared with the target, so comparing an observation
to a user-specified goal~(\pr{1}) requires a separate registration step, and there is no
direct signal for simultaneously decomposing local wrinkle severity and pose
offset~(\pr{2}, \pr{3}).
Template-based tracking~\cite{wang2024trtm} introduces persistent vertex correspondence by continuously deforming a canonical mesh; this work extends that representation to also define the target state, unlocking direct per-vertex error computation
without any registration.

\subsection{IL for manipulation}

Teleoperation-based \ac{IL} avoids explicit dynamics modeling and sim-to-real gaps by learning directly from expert demonstrations.
Action Chunking Transformers (ACT)~\cite{zhao2023learning, zhao2024aloha} condition a Transformer on image observations to predict chunked action sequences, while diffusion-based policies~\cite{black2024pi_0,Ze2024DP3} generate multimodal action distributions. These methods typically rely on RGB or RGB-D observations and large demonstration datasets, limiting alignment precision and generalization to novel target configurations in RTFF. This work builds on ACT~\cite{zhao2023learning} but replaces image inputs with mesh vertex positions and real-time alignment error, enabling goal-conditioned RTFF from a compact demonstration set and making the policy directly sensitive to 3D wrinkle geometry and pose offset~(\pr{2}, \pr{3}).

\section{Proposed framework outline}


\subsection{Robotic manipulation platform}\label{sec:robotic_manipulation_platform}

The robotic manipulation platform (Fig.~\ref{fig:overall_framework}~(b)) consists of two 6‑DoF manipulators suspended from an aluminum frame, each with a wrist‑mounted F/T sensor and a custom end‑effector. Dual‑arm impedance control~\cite{Kai2025} enables safe interaction with the environment without joint torque sensing. An RGB‑D sensor on top provides images for real‑time state estimation. Human demonstration data for RTFF is collected via teleoperation using two SpaceMouse devices to command end‑effector velocities and desired forces. A low‑level workstation provides high-frequency control of the manipulators, while a high‑level computer runs state estimation and RTFF policy modules at a lower frequency.

For operating the platform, a world frame $\Sigma_\mathrm{w}$ (Fig.~\ref{fig:overall_framework}~(b)) is defined with the $z$‑axis normal to the table and all vectors are expressed in this frame. At time $t$, the proprioceptive state of end‑effector $i \in \{L,R\}$ ($L$ left, $R$ right) is $\mathbf{m}_{t,i}\!\in\!\mathbb{R}^{12}$, containing its 6D pose and wrench; the combined state is $\mathbf{m}_t\!=\![\mathbf{m}_{t,L}^\top, \mathbf{m}_{t,R}^\top]^\top \in \mathbb{R}^{24}$. The action of the manipulator $i$ is  
\(\mathbf{a}_{t,i}=[v_{x,i}, v_{y,i}, \omega_{z,i}, F^{d}_{z,i}]^\top\),  
with $v_{x,i}, v_{y,i}$ planar velocities, $\omega_{z,i}$ yaw rate, and $F^{d}_{z,i}$ the desired vertical force for impedance control.  
The joint action $\mathbf{a}_t=[\mathbf{a}_{t,L}^\top, \mathbf{a}_{t,R}^\top]^\top \in \mathbb{R}^8$ drives the end‑effectors for fabric manipulation.

\subsection{RTFF task description}

Let the canonical template mesh be \(\mathcal{T}=(V^{\text{TPL}},E^{\text{TPL}})\), a standardized flattened fabric reference derived from the fabric CAD model. Here, \(V^{\text{TPL}}=\{ \mathbf{v}^{\text{TPL}}_i \}_{i=1}^N\) are vertex positions, and \(E^{\text{TPL}}\subseteq\{(i,j)\mid i\neq j\}\), \(|E^{\text{TPL}}|=N_e\), are triangular mesh that connects the vertices. 
At time $t$, the RGB‑D input and $\mathcal{T}$ are used for state estimation to predict the deformed mesh \(\hat{M}_t=(\hat{V}_t,E^{\text{TPL}})\), with \(\hat{V}_t=\{\hat{\mathbf{v}}_{i,t}\}_{i=1}^N\). Reusing $E^{\text{TPL}}$ preserves vertex connectivity. 
Given a wrinkle‑free, flattened target mesh derived from the template mesh (red, Fig.~\ref{fig:introduction_figure}),
\(
M^*\!=\!(V^*, E^\text{TPL}),
\)
with 
\(
V^*\!=\!\{ \mathbf{v}^*_i \}_{i=1}^N
\) encoding the user-specified target pose (\pr{1}), 
the \textit{mean alignment error} between the predicted mesh $\hat{M}_t$ and the target mesh $M^*$ at time $t$ along $x,y,z$ is
\begin{equation}
    \bar{\mathbf{e}}_t = \frac{1}{N}\sum_{i=1}^N 
    \mathbf{e}_{i,t},\quad\text{where}\quad\mathbf{e}_{i,t} = \hat{\mathbf{v}}_{i,t} - \mathbf{v}^*_i\in\!\mathbb{R}^{3}.
    \label{eq:alignment_vector_error}
\end{equation}
Here, \(\bar{\mathbf{e}}_t \in \mathbb{R}^3\), and \(\mathbf{e}_{i,t}\) denotes the $i^\text{th}$ {vertex alignment error}, providing the per-vertex wrinkle and pose feedback required
by (\pr{3}).  
The template mesh $\mathcal{T}$, shared by $\hat{M}_t$ and $M^*$, ensures correspondence‑preserving state estimation, deformation tracking, and sim‑to‑real transfer.
\textbf{Note:} Throughout the paper, 
by a slight abuse of notation, $M$ (such as $M^*, \hat{M}$) denotes the vertex position \(V\) (such as $V^*, \hat{V}$).

\begin{definition}[RTFF task]
Assuming fabric wrinkles do not involve irreversible deformations (such as stretched fibers or permanent creases) and that the fabric can be fully flattened on the table with a sufficient sequence of dual‑arm actions \(\mathbf{a}_{0:T-1}=\{\mathbf{a}_0,\dots,\mathbf{a}_{T-1}\}\), the RTFF task is formulated as minimizing the terminal mean alignment magnitude (terminal mean alignment norm) over \(\mathbf{a}_{0:T-1}\)   
\begin{equation}
    \min_{\mathbf{a}_{0:T-1}} \; \big\| \bar{\mathbf{e}}_T(M_T,M^*) \big\|_2 ,
    \label{eq:rtff_optimization}
\end{equation}
where $T$ denotes the planning horizon. Equation~\eqref{eq:rtff_optimization} evaluates the final state after \(\mathbf{a}_{0:T-1}\) transforms the initial wrinkled mesh $\hat{M}_0$ into $M^*$.
\end{definition}

\subsection{RTFF policy}
\label{sec:proposed_framework}

The key challenge in \ac{RTFF} is wrinkled fabric state estimation under occluded RGB‑D from the manipulator end-effectors, while tracking mesh states to plan manipulator actions. This section first introduces state estimation to handle occlusions, followed by the policy.
\subsubsection{Fabric state estimation}
\label{sec:fabric_state_estimation}
The fabric state estimation of RTFF policy predicts the fabric mesh (green, Fig.~\ref{fig:introduction_figure}). The current RGB image and
point cloud are first processed by a fabric depth rendering module to produce segmented point cloud $Q^{\text{SEG}}_t$, and segmented fabric depth image \(D^{\text{SEG}}_t\) at time \(t\). Next, \(D^{\text{SEG}}_t\) and \(\mathcal{T}\) are input to a \ac{GAT}, \(\mathcal{G}(\cdot)\), trained on synthetic data from a \textit{triangular mass–spring mesh model} with \textit{occlusions}, to predict the deformed fabric mesh \(\hat{M}^\text{GAT}_t\) in real time. However, to ensure smoother mesh‑prediction transitions between consecutive time steps of the manipulation policy, and further improve accuracy, sim‑to‑real fine‑tuning is performed, in which $\hat{M}^{\text{GAT}}_t$ is refined by a residual correction $\delta \hat{M}\!\left(\hat{M}^{\text{GAT}}, Q^{\text{SEG}}\right)$ by optimizing a one-sided Chamfer loss that aligns the mesh with $Q^{\text{SEG}}$.
The final prediction can be written as 
\begin{equation}
    \label{eq:GAT}
    \textstyle
    \hat{M}_t = \textstyle{\hat{M}_t^\text{GAT}}+\delta \hat{M}_t\!\left(\cdot\right),\text{ }\text{with}\quad{\hat{M}_t^\text{GAT}}=\mathcal{G}(D^{\text{SEG}}_t, \mathcal{T}).
\end{equation}
\subsubsection{Fabric manipulation policy} 
\label{sec:fabric_manipulation_policy}
Given the prediction from \eqref{eq:GAT}, fabric manipulation is formulated as a Markov Decision Process (MDP) 
$\mathcal{M}=(\mathcal{S},\mathcal{A},\mathcal{P})$, where the observation state at time 
$t$ is ${\mathbf{s}}_t=\{{M}^*,\hat{M}_t,D_t,\mathcal{E}_t,\mathbf{m}_t\}\in\mathcal{S}$, 
where $D_t$ is the environment depth image and \textit{alignment error} \(
\mathcal{E}_t = \{ \mathbf{e}_{i,t} \}_{i=1}^N
\). Given ${\mathbf{s}}_t$ and action 
$\mathbf{a}_t\in\mathcal{A}$, transition 
$\mathcal{P}({\mathbf{s}}_{t+1}\mid{\mathbf{s}}_t,\mathbf{a}_t)$ captures nonlinear 
fabric dynamics.

To solve the RTFF task, a hybrid manipulation policy 
$\pi_\theta : \mathcal{S}\!\to\!\mathcal{A}$ is proposed, consisting of an IL 
policy $\pi_{\theta}^{\text{IL}}$ for coarse alignment and a VS policy 
$\pi_{\theta}^{\text{VS}}$ for fine alignment, jointly addressing simultaneous flattening and
pose alignment (\pr{2}). 
Given state $\mathbf{s}^{\text{IL}}_t=\mathbf{s}_t$, the IL policy is modeled as the
conditional distribution
\(
    \pi_{\theta}^{\text{IL}}(\mathbf{a}_t \mid \mathbf{s}^{\text{IL}}_t),
    \label{eq:policy}
\)
which is fit implicitly from human demonstrations. The learned policy generates
sequence of manipulation actions $\mathbf{a}_t$ such as grasp point selection, fabric 
flattening, pose adjustment, and coarse alignment to $M^*$. 
For the VS policy, the reduced state 
$\mathbf{s}^{\text{VS}}_t=\{{M}^*,\hat{M}_t,\mathcal{E}_t,\mathbf{m}_t\}$ 
is used to design a closed-loop control law 
\(
    \pi_{\theta}^{\text{VS}}(\mathbf{s}^{\text{VS}}_t)
\)
which coordinates dual-arm actions for fine alignment. 
Note that $\mathcal{P}$ is implicit in IL (learned from demonstrations) and explicit in VS (defined by Jacobian).

\begin{definition}[RTFF policy]
The hybrid policy is given by 
%
\begin{equation}
\textstyle
\mathbf{a}_t \sim 
\begin{cases}
    \pi_{\theta}^{\text{VS}},  & \text{if } \Phi_t \geq \epsilon_\Phi,\; \|\mathbf{\bar{e}}_t\|_2 \leq \epsilon_{\mathbf{\bar{e}}},\; F_{z,i,t} \geq F_z^g ,
    \\
    \pi_{\theta}^{\text{IL}}, & \text{otherwise}, 
\end{cases}
\label{eq:hybrid_policy}
\end{equation}
%
where $\Phi_t\!\in\![0,1]$ is flatness, fraction of mesh vertices on tabletop within a height tolerance;
$\|\mathbf{\bar{e}}_t\|_2$ is the alignment error magnitude (mean alignment error norm); and
$F_{z,i,t}$ is the end-effector vertical force with $i\!\in\!\{L,R\}$.
The IL policy first performs coarse alignment to reduce wrinkles and bring the fabric closer to $M^*$.
When the fabric is flat and aligned, $\Phi_t\!\geq\!\epsilon_\Phi$ and $\|{\mathbf{\bar{e}}}_t\|_2\!\leq\!\epsilon_{\mathbf{\bar{e}}}$, IL selects grasping points for handoff.
After both arms grasp the fabric, $F_{z,i,t}\!\geq\!F_z^g$, control switches to $\pi_{\theta}^{\text{VS}}$ for fine alignment, producing the action sequence ${\mathbf{a}}_{0:T-1}$ that minimizes \eqref{eq:rtff_optimization}.
\end{definition}

\subsubsection{\ac{MACT}} %
\label{sec:mact}

This work implements \ac{IL} with a novel \ac{MACT} policy conditioned on the predicted mesh \(\hat{M}\) from~\eqref{eq:GAT}, the target mesh \(M^*\), and the alignment error \(\mathcal{E}_t\) using a Transformer backbone. The policy outputs action chunks~\cite{zhao2023learning} for efficient storage and execution while capturing \textit{non-Markovian} behaviors in demonstrations.
Given an observation $\mathbf{s}^{\text{IL}}_t$, \ac{MACT} predicts a horizon of $N_h$ actions $\hat{\mathbf{a}}_{t:t+N_h}$. Hence the one‑step IL policy $\pi^{\text{IL}}_\theta(\mathbf{a}_t \mid \mathbf{s}^{\text{IL}}_t)$ in \eqref{eq:hybrid_policy} can be extended and rewritten as
\(
\pi^{\text{IL}}_\theta(\hat{\mathbf{a}}_{t:t+N_h}\mid \mathbf{s}^{\text{IL}}_t)
\).
At deployment, this horizon is executed in a receding manner via temporal ensembling: overlapping predictions are averaged to yield the next action, thereby recovering $\pi^{\text{IL}}_\theta(\mathbf{a}_t\mid \mathbf{s}^{\text{IL}}_t)$.
Finally, MACT supervision is provided by a reconstruction loss, computed as the $\ell_1$ discrepancy between \(\hat{\mathbf{a}}_{t:t+N_h}\) and the ground‑truth chunk \(\mathbf{a}_{t:t+N_h}\) from demonstrations, averaged across $N_h$. 

\section{Methodology}
\label{sec:methodology}
%
\subsection{State estimation implementation}

\subsubsection{Synthetic data generation}
\label{sec:synthetic_data}

A fabric state is modeled as a triangular mass--spring mesh
\(
M^{\text{SIM}} = (V^{\text{SIM}}, E^{\text{TPL}}),
\)
where $V^{\text{SIM}}\!=\!\{\mathbf{v}^\text{SIM}_i\}_{i=1}^N$ are 3D vertex positions and $E^\text{TPL}$ connects neighboring vertices as point masses. Each edge $(i,j)\!\in\!E^{\text{TPL}}$ acts as a spring with vector 
\(
\mathbf{d}^{\text{SIM}}_{ij} = \mathbf{v}^{\text{SIM}}_j - \mathbf{v}^{\text{SIM}}_i, 
\quad \|\mathbf{d}^{\text{SIM}}_{ij}\|_2,
\)
enforcing structural constraints. This mesh is implemented in \textit{Blender} \cite{blender2025}.  
To generate diverse states, randomized manipulations were applied, including multi‑hand and single‑hand interactions (with or without lifting), as well as compression and shearing. After each manipulation, the final deformed mesh ${M}^{\text{SIM}}$ is saved as the simulation output.

After mesh generation, depth images are rendered with \textit{PyTorch3D} \cite{ravi2020accelerating}. Centralized meshes are batched, processed with a custom depth shader, and z-buffer values normalized to a fixed range. The outputs are converted to single-channel grayscale depth images, and the resulting $D^{\text{SIM}}$ from $M^{\text{SIM}}$ are saved for GAT training. Finally, each $D^{\text{SIM}}$ is converted to a simulated point cloud
\(
Q^{\text{SIM}}=\{\mathbf{q}^{\text{SIM}}_j\in\mathbb{R}^3 \mid j=1,\dots,N^{\text{SIM}}_q\},
\)
by mapping each pixel to a 3D point.

\subsubsection{Mesh prediction model}
\label{sec:state_estimation}
The model follows a GAT $\mathcal{G}(\cdot)$ framework as discussed in \cite{wang2024trtm}, supervised on synthetic datasets (Sec.~\ref{sec:synthetic_data}).  
The objective is  
\begin{align}
\mathcal{L}_{\text{GAT-Train}} = \mathcal{L}_{\text{Vtx}} + \lambda_k \mathcal{L}_{\text{Key}} + \lambda_c \mathcal{L}_{\text{Cham}} .
\label{eq:train_loss}
\end{align}
The vertex loss measures $\ell_1$ distance between predicted vertices $\hat{\mathbf{v}}^{\text{GAT}}_i$ and \ac{GT} ${\mathbf{v}}^\text{SIM}_i$,  
while the keypoint loss applies the same penalty on semantic keypoints (neckline, shoulder, arm, hem) 
\begin{align}
\textstyle
\mathcal{L}_{\text{Vtx}} = \tfrac{1}{N} \sum_{i=1}^{N} \lVert \Delta \mathbf{v}_i \rVert_1, \hspace{0.5em}
\mathcal{L}_{\text{Key}} = \tfrac{1}{N_k} \sum_{i \in \text{Key}} \lVert \Delta \mathbf{v}_i \rVert_1 ,
\label{eq:loss_vertex}
\end{align}
where $\Delta \mathbf{v}_i = \hat{\mathbf{v}}^{\text{GAT}}_i - \mathbf{v}^\text{SIM}_i$, \(N_k\) denotes number of keypoints.
To refine boundary and surface similarity, the Chamfer loss measures distances from the simulated point cloud $Q^{\text{SIM}}$ to predicted vertices 
\begin{equation}
\textstyle
\mathcal{L}_{\text{Cham}} = \tfrac{1}{N_q^{\text{SIM}}} \sum_{{\mathbf q}^\text{SIM}_j\in Q^{\text{SIM}} } \min_{{\mathbf{\hat v}}^{\text{GAT}}_i } \lVert {\mathbf{\hat v}}^{\text{GAT}}_i - {\mathbf{q}}^\text{SIM}_j \rVert_2 .
\label{eq:loss_chamfer}
\end{equation}
Finally, $\mathcal{G}(\cdot)$ is trained by minimizing \eqref{eq:train_loss} with meshes $M^{\text{SIM}}$ as ground truth and depth images $D^{\text{SIM}}$ as input.

\subsubsection{Real fabric depth rendering}
\label{sec:real_fabric_depth}
At time $t$, the segmented depth image $D^{\text{SEG}}_t$ is produced by a GPU-accelerated pipeline.  
RGB and point cloud data are first captured with the RealSense sensor (Sec.~\ref{sec:robotic_manipulation_platform}).  
\textit{SAM2}~\cite{ravi2024sam2} segments the RGB frame to obtain a fabric mask, which is applied to the point cloud using Torch functions to isolate fabric points.  
The resulting cloud is denoised and centralized, yielding
\(
Q^{\text{SEG}}_t = \{ \mathbf{q}^{\text{SEG}}_j \in \mathbb{R}^3 \mid j=1,\dots,N_{q}^{\text{SEG}} \},
\)
where $N_{q,t}^{\text{SEG}}$ is the number of points.  
Finally, $Q^{\text{SEG}}_t$ is rasterized with PyTorch3D to form $D^{\text{SEG}}_t$ for fabric state estimation.

\subsubsection{Sim-to-real fine-tuning}

To obtain the residual correction 
$\delta \hat{M}\!\left(\hat{M}^{\text{GAT}}, Q^{\text{SEG}}\right)$ defined in Sec. \ref{sec:fabric_state_estimation}, a one-sided Chamfer loss that aligns the GAT predicted mesh $\hat{M}^{\text{GAT}}$ with the segmented point cloud $Q^{\text{SEG}}$ is written as
\begin{equation}
\textstyle
\mathcal{L}_{\text{Cham-R}} =
\frac{1}{N_q^{\text{SEG}}} \sum_{\mathbf{q}_j^{\text{SEG}} \in Q^{\text{SEG}}}
\min_{\hat{\mathbf{v}}_i^{\text{GAT}}}
\|\hat{\mathbf{v}}_i^{\text{GAT}} - \mathbf{q}^{\text{SEG}}_j \|_2 .
\label{eq:chamfer_loss_real}
\end{equation}
To avoid vertex drift, an edge-length regularization preserves local mesh structure is defined as
\begin{align}
\textstyle
\mathcal{L}_{\text{Edge-R}} = \tfrac{1}{N_e} \sum_{(i,j)\in E^\text{TPL}} 
\big| \lVert {\mathbf{\hat d}}^{\text{GAT}}_{ij}\rVert_2 - \lVert \mathbf{d}_{ij}^\text{TPL}\rVert_2 \big| .
\label{eq:loss_regularization_real}
\end{align}
The final objective is 
\begin{equation}
\textstyle
\mathcal{L}_\text{R} =
\lambda_{\text{Cham-R}} \, \mathcal{L}_{\text{Cham-R}}
+ \lambda_{\text{Edge-R}} \, \mathcal{L}_{\text{Edge-R}},
\label{eq:optimization}
\end{equation}
where weights $\lambda_{\text{Cham-R}}$ and $\lambda_{\text{Edge-R}}$ balance alignment and regularity.
Minimization of $\mathcal{L}_\text{R}$ provides the residual correction \(\delta \hat{M}\) to the GAT prediction, defined as
\(
\delta \hat{M} \;=\; 
\arg\min_{\Delta M} \;
\mathcal{L}_\text{R}\!\left(\hat{M}^{\text{GAT}} + \Delta M, Q^{\text{SEG}}\right),
\)
which yields the final predicted mesh \eqref{eq:GAT}. 

\subsection{Hybrid manipulation policy implementation}

\subsubsection{\acf{MACT}}
\begin{figure}
    \centering
    \includegraphics[width=1\linewidth]{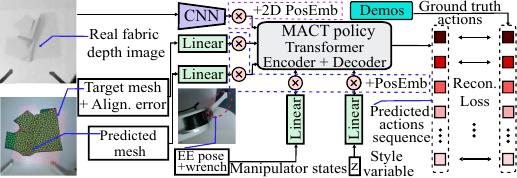}
    \caption{MACT policy architecture.}
    \label{fig:ACT}
    \vspace{-1.7em}
\end{figure}

The MACT policy (Sec. \ref{sec:fabric_manipulation_policy}, Fig. \ref{fig:ACT}) takes as input the raw environment depth image $D_t\!\in\!\mathbb{R}^{H\!\times\!W\!\times 1}$, which is encoded by a ResNet‑18 backbone into a feature map of size $\mathbb{R}^{H'\!\times W'\!\times\!d_\text{model}}$, where $d_\text{model}$ is a predefined model dimension. The feature is then flattened along the spatial dimension to obtain a $(H'W')\times d_\text{model}$ token sequence. To preserve spatial information, a 2D sinusoidal positional embedding is added to the feature sequence.  
Additional inputs include: (i) the target mesh $V^*$ with alignment error, $M^* \oplus \mathcal{E}_t \in \mathbb{R}^{N \times 6}$, embedded to $N\times d_\text{model}$; (ii) the predicted mesh vertices $\hat{V} \in \mathbb{R}^{N \times 3}$,  embedded to $N\times d_\text{model}$; (iii) the EE states $\mathbf{m}_t\!=\![\mathbf{m}^\top_{t,L}, \mathbf{m}^\top_{t,R}]^\top\!\in\!\mathbb{R}^{1 \times 24}$ ({i.e., } two manipulators, each with 6D position + 6D wrench), embedded to $1\times d_\text{model}$. Additional sinusoidal positional embedding is added to these input sequences.

A style variable $z\!\in\!\mathbb{R}^{1 \times 24}$ is output by Transformer‑based Conditional Variational Autoencoder (CVAE).  
It takes as input the EE states and a target action sequence of length $N_h$ prepended with a learned “[CLS]” token, forming a $(N_h+2)$ sequence.  
The “[CLS]” feature outputs the mean and variance of $z$, which is sampled for training and set to the prior mean (zero) at inference for deterministic decoding.  
$z$ is projected to $1 \times d_\text{model}$ via linear layers, yielding the concatenated encoder input ${\mathbf{s}}_t^\text{IL} \in \mathbb{R}^{(H'W' + 2N + 2) \times d_\text{model}}$.

The MACT decoder attends to encoder features via cross‑attention, using a fixed $N_h \times d_\text{model}$ positional sequence as input.  
It outputs an $N_h \times d_\text{model}$ sequence, down‑projected by an MLP to $N_h \times 8$, representing manipulator actions  
$\mathbf{a}_t = [\mathbf{a}^\top_{t,L}, \mathbf{a}^\top_{t,R}]^\top \in \mathbb{R}^8$  
for the two manipulators over $N_h$ steps.  
The model is trained with a VAE objective: reconstruction of action chunks plus a regularization term aligning the VAE encoder with a Gaussian prior.

\subsubsection{\acf{VS}}
%
%
At time $t$, the predicted mesh 
$\hat{M}_t\!=\!(\hat{V}_t,E^{\text{TPL}})$ meets the condition of $\pi_{\theta}^{\text{VS}}$ in~\eqref{eq:hybrid_policy} to switch to VS. 
Assuming the fabric undergoes a rigid-body motion under dual-arm control~\cite{tokuda2025transformer}, the mean alignment error between $\hat{M}_t$ and the target mesh $M^*=(V^*,E^{\text{TPL}})$ given by~\eqref{eq:alignment_vector_error} is reformulated as an objective function
\begin{equation}
    \textstyle
    \min_{R_t \in SO(3),\, \mathbf{t}_t \in \mathbb{R}^3} 
    \frac{1}{N}\sum_{i=1}^N \big\| (R_t \hat{\mathbf{v}}_{i,t} + \mathbf{t}_t) -\mathbf{v}_i^*\big\|^2,
\end{equation}
where $\hat{\mathbf{v}}_{i,t}\!\in\!\hat{V}_t$, $\mathbf{v}_i^*\!\in\!V^*$ are corresponding vertices, and $T_t=(R_t,\mathbf{t}_t)\in SE(3)$ is the rigid-body transformation matrix w.r.t.\ $\Sigma_\mathrm{w}$.  
Since motion is restricted to the tabletop, $T_t$ reduces to planar translation $\mathbf{p}_t=[x_t,y_t]^\top$ and rotation $\theta_t=\operatorname{atan2}(R_t(2,1),R_t(1,1))$.  
The pose error vector is then $\mathbf{\bar{e}}_t^p=(\mathbf{p}_t^\top,\theta_t)$, where $\mathbf{p}_t$ is the residual translation, and $\theta_t$ is the rotation to align $\hat{M}_t$ with $M^*$.

To eliminate this \(
{\mathbf{\bar{e}}}_t^p
\), a coordinated motion of the dual-arm manipulator is calculated by introducing position-based VS and rigid-body motion. The control motion is a planar spatial twist, \(\mathbf{a}_t^{vs} = [\,v_x,\, v_y,\, \omega_z\,]^\top\subset\mathbf{a}_t\), defined in Sec.~\ref{sec:robotic_manipulation_platform}, and
$F_{z}^d$ is set as a constant.
In position-based VS, the error dynamics are written as
$
\dot{\bar{\mathbf{e}}}_t^p = L \, \mathbf{a}_t^{vs},
$
where $L$ is the interaction matrix (Jacobian) by assuming a rigid body motion at time $t$. By imposing exponential 
decay of the error as $\dot{\bar{{\mathbf{e}}}}_t^p = -\lambda \mathbf{\bar{e}}_t^p$ with a 
gain $\lambda \in\mathbb{R}$, the control input becomes
\begin{equation}
    \mathbf{a}_t^{vs} = L^\dagger\dot{\bar{\mathbf{e}}}_t^p=-\lambda L^\dagger \mathbf{\bar{e}}_t^p,
\end{equation}
where $L^\dagger$ is the Moore–Penrose pseudo-inverse of $L$. The control input is calculated at each iteration, providing real-time feedback that mitigates the impact of rigid-body assumptions and increases alignment precision~\cite{Edmund2025}.

\section{Results}

\subsection{Hardware specifications}
The manipulators in the proposed platform (Fig.~\ref{fig:overall_framework}~(b)) are DENSO VS-068 arms, each with an ATI Axia80-M8 F/T sensor and a custom 3D‑printed end-effector. Teleoperation uses two SpaceMouse\textsuperscript{\textregistered} Compact devices.
The vision system is an Intel\textsuperscript{\textregistered} RealSense\texttrademark{} LiDAR Camera L515. The lower-level workstation is powered by an Intel\textsuperscript{\textregistered} Xeon\textsuperscript{\textregistered} W-2295 CPU running INtime OS over EtherCAT at 4 kHz, and the high-level computer is equipped with an NVIDIA GeForce RTX~4090 GPU, an Intel\textsuperscript{\textregistered} Core\texttrademark{} i9-14900KF CPU, and 128\,GB RAM running Ubuntu~24.04 at 20 Hz.

\subsection{State estimation results}
\begin{figure}
    \centering
    \includegraphics[width=1\linewidth]{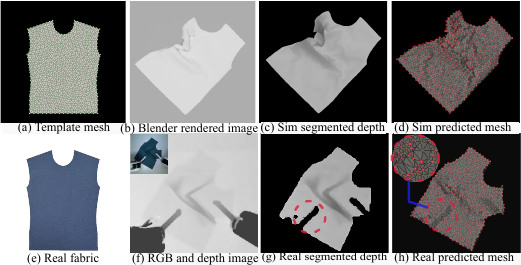}
    \caption{Simulated and real fabric mesh prediction.}
    \label{fig:state_estimation}
    \vspace{-1.7em}
\end{figure}

This section presents state estimation results on a \textit{T-shirt front panel} fabric with dimensions 400$\times$550~mm (Fig.~\ref{fig:state_estimation} (e)). The canonical template mesh $\mathcal{T}$, with $N\!=\!562$ vertices (Fig.~\ref{fig:state_estimation}~(a)), is generated in MeshLab~\cite{Meshlab}.

\paragraph*{Synthetic data generation}
For synthetic data generation in Blender, fabric simulation parameters were manually tuned for real fabric behavior. 
Random perturbations are applied during data generation to account for uncertain fabric parameters and mitigate the sim‑to‑real gap. Rendered cloth example in Blender is shown in Fig. \ref{fig:state_estimation}~(b).
Finally, to generate simulated segmented depth image Fig. \ref{fig:state_estimation}~(c) for GAT training, the simulated fabric meshes $M_t$ are saved and processed through the same depth rendering pipeline (Sec. \ref{sec:real_fabric_depth}) as real data to ensure consistency  (Fig.~\ref{fig:state_estimation}~(c)).

\paragraph*{Training mesh prediction model}
About 60{,}000 simulated depth--mesh pairs were generated and split into 80\% training, 10\% validation, and 10\% testing to train the GAT (Sec.~\ref{sec:state_estimation}) for wrinkle-aware state estimation~(\pr{3}) in the RTFF policy. Training used $512\times512$ depth inputs, batch size 16, 500 epochs, and learning rate $1\times10^{-4}$; the loss weights in \eqref{eq:train_loss} were $\lambda_k=1$ and $\lambda_c=0.5$. Data augmentation included: (i) additive Gaussian noise; (ii) random rigid-body rotations; (iii) synthetic occlusions of various shapes and sizes to mimic EE and manipulator (external) occlusions; and (iv) pixel-level $z$-variations to capture wrinkles. For sim-to-real, the mesh $\hat{M}^{\text{GAT}}$ is predicted from segmented depth $D^{\text{SEG}}$ using the trained GAT $\mathcal{G}(\cdot)$, then refined in real time by optimizing the one-sided Chamfer loss \eqref{eq:chamfer_loss_real} with Adam using $\lambda_{\text{Cham-R}}=1$, $\lambda_{\text{Edge-R}}=0.01$, learning rate $2\times10^{-3}$, and five steps per observation to yield the final predicted mesh $\hat{M}$.

\begin{figure}
    \centering
    \includegraphics[width=1\linewidth]{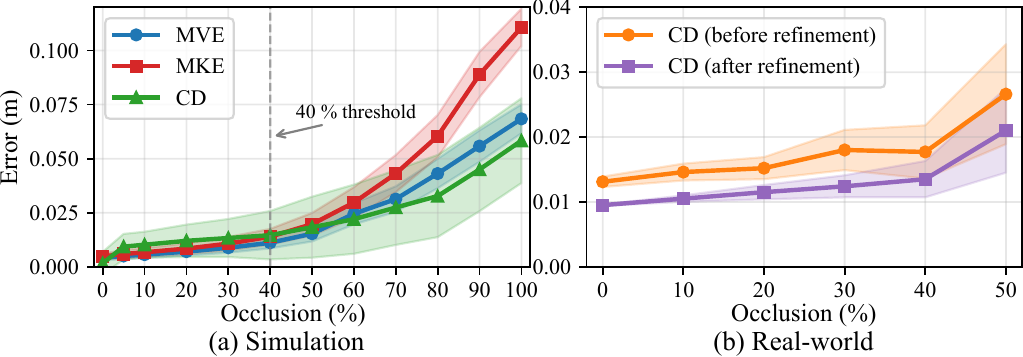}
    \caption{Evaluation results with controlled occlusion.}
    \label{fig:state_estimation_graph}
    \vspace{-1.7em}
\end{figure}

\paragraph*{Evaluation}
State estimation is first shown in simulation, where depth (Fig.~\ref{fig:state_estimation}~(c)) is applied to the trained GAT to predict meshes that recover fabric state (Fig.~\ref{fig:state_estimation}~(d)). 
For evaluation, the trained GAT was tested on a held-out test set under synthetically masked occlusion, reporting \textit{Mean Vertex Error (MVE) and Mean Keypoint Error (MKE)} in Fig.~\ref{fig:state_estimation_graph}.
Here, {MVE} and {MKE} are the mean Euclidean distances between predicted and \ac{GT} mesh vertices and semantic keypoints (Sec.~\ref{sec:state_estimation}), respectively.
The network maintains high accuracy under moderate occlusion (MVE = 0.0070 m at 20\%, 0.0111 m at 40\%), but errors accumulate steeply beyond 40\%, establishing a practical guideline to keep occlusion below 40\%.

In real-world experiments, regions missing due to EE occlusion (Fig.~\ref{fig:state_estimation}(f)--(g), red circle) are inferred using the GAT prediction, with Chamfer loss-based refinement improving accuracy (Fig.~\ref{fig:state_estimation}(h)).
For evaluation, where \ac{GT} is unavailable, \textit{One-sided Chamfer Distance} (OCD) from the predicted mesh to the detected depth image was measured under controlled occlusion. Initial predictions showed slightly higher OCD than simulation, but refinement reduced OCD by 21--31\% across all occlusion levels, aligning with simulation performance and confirming accurate mesh prediction in the nominal regime.

\subsection{Manipulation policy results}

This section demonstrates the performance of the proposed RTFF manipulation policy pipeline. The policy is built on the LeRobot framework~\cite{cadene2024lerobot} and trained with human demonstrations collected on the robotic platform discussed in Sec.~\ref{sec:robotic_manipulation_platform}. 
The system runs across several parallel threads: the main MACT control thread at 20 Hz and the mesh prediction thread at 10 Hz, where SAM2 segmentation, depth rendering, and GAT forward pass account for 30\%, 35\%, and 20\% of computation, respectively.

During training, demonstrations are collected using a fabric panel of thickness 0.32~mm, areal density 0.21~kg/m$^2$, and Young’s modulus 21.81~MPa. 
Because fabric placement and wrinkle formation are inherently stochastic on this platform, identical states cannot be reproduced. Thus, the policy must be strictly observation‑driven for localizing wrinkles, planning motions to flatten and rotate the fabric to the target (\pr{1}, \pr{2}), selecting grasping points before switching to VS using the current and target meshes (\pr{3}), and coordinating collision‑free dual-arm actions.

Finally, the data collection and evaluation process on this platform proceeds as follows:  
(i) the dual-arm manipulator is initialized to a home pose, and a wrinkled fabric is placed on the table;  
(ii) a user-specified target state $M^*$ (\pr{1}) is randomly generated within the camera’s field of view;  
(iii) demonstrations are collected by manually controlling the manipulators with a SpaceMouse to perform RTFF across multiple episodes, restricted to a single fabric type with fixed material and scale;  
(iv) these demonstrations are used to train the \ac{MACT} component of the RTFF policy;  
(v) the trained policy is executed on fabrics with previously unseen wrinkles, thickness, textures, materials, and scales to jointly remove all wrinkles and flatten them (\pr{2}) toward the target state;  
(vi) finally, the manipulator returns to the home pose and experimental results are recorded for evaluation.

\subsubsection{Experimental protocol and evaluation metrics} 

\begin{figure*}[t!]
    \centering
    \includegraphics[width=0.995\textwidth]{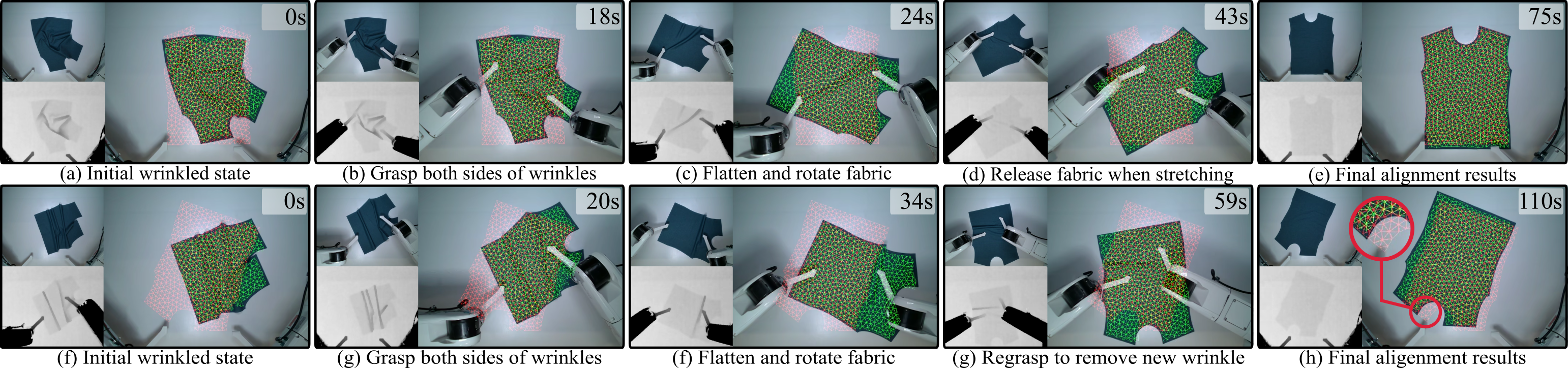}
    \caption{Example RTFF sequences with (top) and without (bottom) Visual Servoing (VS).}
    \label{fig:timeseries}
    \vspace{-1.7em}
\end{figure*}

For experiments, a pre‑defined set of fabric states and targets was needed, but consistent wrinkle reproduction was infeasible. To ensure fair comparison, 20 target meshes $M^*$ were pre‑sampled and used in all trials. Initial states were generated by orienting the neckline at 0°, 90°, 180°, and 270°, each with five trials introducing random wrinkles, yielding 20 states matching the 20 targets. Experimental parameters were set to $\epsilon_{\mathbf{\bar{e}}}=0.05\,\text{m}$, $\epsilon_\Phi=0.95$, and $F_z^g=2.5\,\text{N}$.
These thresholds are empirically selected to ensure \ac{VS} is triggered only when the fabric is sufficiently flat, the alignment error is within the convergence range of \ac{VS}, and the fabric is securely grasped by both end-effectors. Switching too early risks unstable servoing on a partially wrinkled fabric, while switching too late wastes manipulation time unnecessarily.

The RTFF policy is evaluated using three metrics:  
(i) \textit{terminal alignment error magnitude (terminal mean alignment error norm)} $\|{\mathbf{\bar{e}}}_T\|_2$, where $\mathbf{\bar{e}}_T$ is given by~\eqref{eq:alignment_vector_error} at $t=T$;  
(ii) \textit{failure}, defined as $\|{\mathbf{\bar{e}}}_T\|_2 > 0.05\,\text{m}$; and  
(iii) \textit{Intersection over Union (IoU)} between the final mesh and target mesh $M^*$, measuring geometric overlap  
\begin{equation}
\text{IoU} = 
\frac{\left| \mathbf{A}_{\text{final}} \cap \mathbf{A}_{M^*} \right|}
     {\left| \mathbf{A}_{\text{final}} \cup \mathbf{A}_{M^*} \right|},
\end{equation}
where $\mathbf{A}_{\text{final}}$ and $\mathbf{A}_{M^*}$ denote the final and target mesh regions. An $\text{IoU}$ of $1$ with sufficiently small $\|{\mathbf{\bar{e}}}_T\|_2$ indicates perfect alignment, while values near $0$ imply negligible overlap. IoU is computed only for non‑failure trials, and experiments terminate early if the first five trials all fail.

\subsubsection{MACT training}
Human demonstrations trained the \ac{IL} (MACT) part of the policy. In total, 50 episodes of 30--90 seconds each were collected at 20 Hz. To broaden the data distribution, random translations were applied to the fabric's initial position. For orientation coverage, the fabric's neckline was uniformly rotated on the table before wrinkles were introduced.
To enhance robustness, online perturbations were introduced: (i) flattened fabric regions were randomly re-wrinkled, and (ii) random displacements from the correct grasping point were applied by the operator via direct end-effector impedance control (without the SpaceMouse).

The MACT component of the RTFF policy was trained with depth inputs ($640\times480$), chunk size $N_h=100$, and model dimension $d_\text{model}=512$. Training used 100{,}000 steps, a batch size of 4, and a learning rate of $1\times10^{-5}$.

\subsubsection{RTFF policy evaluation} 

Table~\ref{table:e1_result}, Row~1, reports the full RTFF policy, showing stable convergence to an average error of about 0.01~m after 20 trials with low variance, outperforming all ablations.
\begin{table}[h!]
\vspace{-0.5em}
\centering
\caption{Ablation study of RTFF policy.}
\resizebox{0.48\textwidth}{!}{%
\begin{tabular}{lccccc}
\toprule
\multirow{2}{*}{Method} & \multicolumn{2}{c}{\textbf{Terminal alignment error (m)}} & \multicolumn{2}{c}{\textbf{IoU}} & \multirow{2}{*}{\textbf{Fail.}} \\ \cline{2-5} 
          & Mean   & Std.    & Mean   & Std.   &      \\
\midrule 
\textbf{Ours}      & \textbf{0.011}  & \textbf{0.002}  & \textbf{0.926} & \textbf{0.018} & \textbf{-}    \\
w/o VS (i)    & 0.028  & 0.010  & 0.871 & 0.05  & -    \\
w/o $\hat{M_t}$ and $\mathcal{E}_t$ (ii) & -      & -      & -      & -     & 5/5  \\
w $D^\text{SEG}_t $, w/o $\hat{M_t}$ and $\mathcal{E}_t$ (iii) & -     & -      & -      &-     & 5/5  \\
\midrule
Type  & 0.0145      &  0.004     & 0.911      & 0.031     & -  \\
Scale  & 0.021      & 0.004      & 0.893      & 0.022     & -  \\
\bottomrule
\end{tabular}%
}
\label{table:e1_result}
\end{table}

During execution (Fig.~\ref{fig:timeseries}, top row), the RTFF policy showed consistent behaviors: 
(i) adjusting the EE pose in the air and manipulating only after a successful grasp;   
(ii) grasping flat areas on both sides of wrinkles and removing them with opposing motions;  
(iii) releasing stretched fabric to restore shape before re-grasping;  
(iv) smoothly switching to VS once criteria given by \eqref{eq:hybrid_policy} are met; and  
(v) switching back to MACT if VS creates new wrinkles and voids \eqref{eq:hybrid_policy}, removing them before resuming VS. This can arise from position--force control coupling effects that cause unexpected end-effector motion during VS execution.
The results show that the policy achieves RTFF by learning fabric state and its relation to the target mesh. Using depth images with a mesh-based representation supports robust wrinkle extraction and accurate displacement estimation between current and target states. With real-time visual feedback, the policy converges reliably across diverse initial configurations and unseen targets.

\subsubsection{Ablation studies}  
Ablations were performed on key RTFF components and MACT.  
The ablation conditions are:  
(i) \textit{Without (w/o) VS:} The IL policy runs for up to 90~s per trial, after which the episode is terminated and outcomes recorded.  
(ii) \textit{W/o current predicted fabric mesh $\hat{M_t}$ and alignment error $\mathcal{E}_t$ embeddings:} Retrain the MACT (Fig.~\ref{fig:ACT}) without $\hat{M_t}$ and $\mathcal{E}_t$ embeddings, with the depth image \(D_t\), robot states and the target mesh.
(iii) \textit{With (w) segmented fabric depth \(D^\text{SEG}_t\), and w/o $\hat{M_t}$ and $\mathcal{E}_t$:} Retrain the MACT with the \(D^\text{SEG}_t\), $D_t$, robot states, and the target mesh as input.

\paragraph*{Ablation study results}
Rows~2--4 of Table~\ref{table:e1_result} report the results, with main findings:  
(i) \textit{W/o VS:} The policy avoids outright failures but shows larger errors and variance, reflecting unreliability (Fig.~\ref{fig:timeseries}, bottom row). Near the target, it fails to extract salient cues, leading to stationary behavior or repeated grasping at the same spot without moving the fabric. 
(ii) \textit{W/o $\hat{M}_t$ and $\mathcal{E}_t$, and (iii) w $D^\text{SEG}_t$:} Both variants fail. Without the mesh-based state estimate and error signal, MACT lacks explicit correspondence between the current and desired fabric states. Replacing them with $D^\text{SEG}_t$ provides geometrically ambiguous observations that the policy cannot exploit from limited demonstrations. In both cases, the policy occasionally flattens local wrinkles but fails to rotate or translate the fabric to match the target configuration.

These results verify the central role of the mesh in the hybrid IL–VS policy. The mesh representation enables sample-efficient IL from few demonstrations, strengthens IL by providing a structured link between the current fabric state and the target, allowing for a smooth switching between IL and VS, and yields a well-defined representation (\pr{3}) that enables VS to converge $\hat{M_t}$ to the given target (\pr{1}).

\subsubsection{Generalization experiments}

\begin{figure}
    \centering
    \includegraphics[width=1\linewidth]{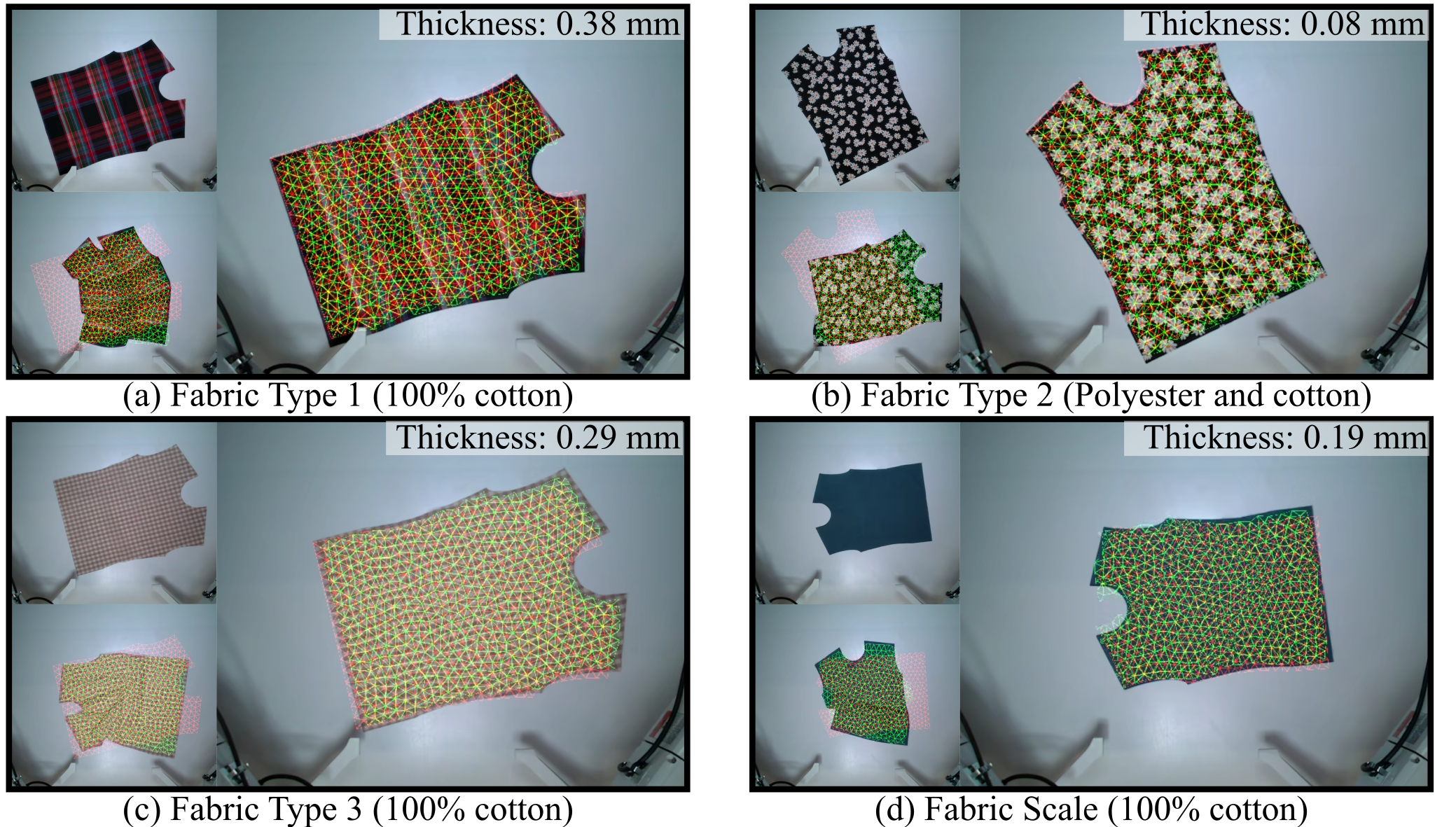}
    \caption{Examples of fabric Types (a)--(c) and Scale (d).}
    \label{fig:generalization}
    \vspace{-1.7em}
\end{figure}

In garment production, fabrics differ in color, material, and size. To test policy generalization, two additional experiments were conducted:  
(i) \textit{Type} — five fabrics of different materials (cotton and polyester) cut to the baseline shape,
with thickness ranging from 0.23--0.40~mm, areal density from 0.14--0.23~kg/m$^2$, and Young's Modulus from 12.43--21.81~MPa. 
(ii) \textit{Scale} — the baseline fabric reduced to 80\% size.  
Each experiment comprised 20 trials, with results given in Rows 5--6 of Table~\ref{table:e1_result} and examples in Fig.~\ref{fig:generalization}.

The results show that the policy, trained solely on the baseline fabric, successfully aligns~(\pr{2}) unseen fabric types and scales to desired user-specified target states~(\pr{1}) without retraining. Although alignment accuracy moderately degrades relative to the baseline, the policy maintains functional performance across all tested configurations. Depth images with predicted mesh enable robust wrinkle features~(\pr{3}) and accurate alignment error estimation between current and target states. Combined with real-time visual feedback, the policy consistently converges across diverse fabric initial states and user-specified target states, indicating strong generalization to varying fabric types and scales.

\acresetall
\section{Conclusion, limitations and future work}

This work presents a \acf{RTFF} policy using a template mesh that anchors both the current fabric state and the user-specified target configuration to the same canonical mesh derived from the fabric’s CAD model. 
This shared reference enables arbitrary target pose specification~(\pr{1}), per-vertex wrinkle state and global pose feedback, and direct inter-mesh error computation without registration~(\pr{3}). Building on this, a template-mesh two-stage \ac{RTFF} policy is proposed~(\pr{2}): a \acf{MACT} \ac{IL} stage learns from a compact demonstration set for coarse alignment and wrinkle reduction guided by mesh state and alignment error, followed by a \ac{VS} stage that drives precise alignment to the target.
Ablation studies confirm the central role of the mesh in the hybrid \ac{IL}--\ac{VS} policy, with each component necessary for reliable \ac{RTFF}, and physical experiments demonstrate strong generalization to unseen target poses, fabric types, and scales.

\subsubsection*{Limitations and future work}
Despite promising results, the framework has several limitations. The system cannot handle folded configurations that require 3D manipulation, and the current policy is shape-specific. Future work will extend the method to support multiple templates for different fabric geometries, improve the inference speed of fabric state estimation, and extend the RTFF policy to full 3D manipulation.

\bibliographystyle{IEEEtran}
\bibliography{Bib/bibliography_new}

\begin{thebibliography}{10}
\providecommand{\url}[1]{#1}
\csname url@samestyle\endcsname
\providecommand{\newblock}{\relax}
\providecommand{\bibinfo}[2]{#2}
\providecommand{\BIBentrySTDinterwordspacing}{\spaceskip=0pt\relax}
\providecommand{\BIBentryALTinterwordstretchfactor}{4}
\providecommand{\BIBentryALTinterwordspacing}{\spaceskip=\fontdimen2\font plus
\BIBentryALTinterwordstretchfactor\fontdimen3\font minus \fontdimen4\font\relax}
\providecommand{\BIBforeignlanguage}[2]{{%
\expandafter\ifx\csname l@#1\endcsname\relax
\typeout{** WARNING: IEEEtran.bst: No hyphenation pattern has been}%
\typeout{** loaded for the language `#1'. Using the pattern for}%
\typeout{** the default language instead.}%
\else
\language=\csname l@#1\endcsname
\fi
#2}}
\providecommand{\BIBdecl}{\relax}
\BIBdecl

\bibitem{Kai2025}
K.~Tang, X.~Huang, A.~Seino, F.~Tokuda, A.~Kobayashi, N.~C. Tien, and K.~Kosuge, ``Fixture-free automated sewing system using dual-arm manipulator and high-speed fabric edge detection,'' \emph{IEEE Robot. Autom. Lett.}, vol.~10, no.~9, pp. 8962--8969, 2025.

\bibitem{taylor1997automated}
P.~Taylor and D.~Pollet, ``Why is automated garment manufacture so difficult?'' in \emph{1997 8th International Conference on Advanced Robotics. Proceedings. ICAR'97}.\hskip 1em plus 0.5em minus 0.4em\relax IEEE, 1997, pp. 39--44.

\bibitem{GRIES2018179}
T.~Gries and V.~Lutz, ``8 - application of robotics in garment manufacturing,'' in \emph{Automation in Garment Manufacturing}, ser. The Textile Institute Book Series.\hskip 1em plus 0.5em minus 0.4em\relax Woodhead Publishing, 2018, pp. 179--197.

\bibitem{Seita2020}
D.~Seita, A.~Ganapathi, R.~Hoque, M.~Hwang, E.~Cen, A.~K. Tanwani, A.~Balakrishna, B.~Thananjeyan, J.~Ichnowski, N.~Jamali, K.~Yamane, S.~Iba, J.~Canny, and K.~Goldberg, ``Deep imitation learning of sequential fabric smoothing from an algorithmic supervisor,'' in \emph{Proc. IEEE/RSJ Int. Conf. Intell. Robots. Syst.}, 2020, pp. 9651--9658.

\bibitem{hietala2022learning}
J.~Hietala, D.~Blanco-Mulero, G.~Alcan, and V.~Kyrki, ``Learning visual feedback control for dynamic cloth folding,'' in \emph{Proc. IEEE/RSJ Int. Conf. Intell. Robots. Syst.}\hskip 1em plus 0.5em minus 0.4em\relax IEEE, 2022, pp. 1455--1462.

\bibitem{tokuda2023cnn}
F.~Tokuda, A.~Seino, A.~Kobayashi, and K.~Kosuge, ``Cnn-based visual servoing for simultaneous positioning and flattening of soft fabric parts,'' in \emph{Proc. IEEE Int. Conf. Robot. Autom.}\hskip 1em plus 0.5em minus 0.4em\relax IEEE, 2023, pp. 748--754.

\bibitem{tokuda2025transformer}
F.~Tokuda, A.~Seino, A.~Kobayashi, K.~Tang, and K.~Kosuge, ``Transformer driven visual servoing for fabric texture matching using dual-arm manipulator,'' \emph{IEEE Robotics and Automation Letters}, vol.~11, no.~2, pp. 1522--1529, 2026.

\bibitem{shetab2023lattice}
M.~Shetab-Bushehri, M.~Aranda, Y.~Mezouar, and E.~{\"O}zg{\"u}r, ``Lattice-based shape tracking and servoing of elastic objects,'' \emph{IEEE Trans. Robot.}, vol.~40, pp. 364--381, 2023.

\bibitem{tang2022track}
T.~Tang and M.~Tomizuka, ``Track deformable objects from point clouds with structure preserved registration,'' \emph{Int. J. Robot. Res.}, vol.~41, no.~6, pp. 599--614, 2022.

\bibitem{lin2021learning}
X.~Lin, Y.~Wang, and D.~Held, ``Learning visible connectivity dynamics for cloth smoothing,'' in \emph{CoRL}, 2021.

\bibitem{huang2023self}
Z.~Huang, X.~Lin, and D.~Held, ``Self-supervised cloth reconstruction via action-conditioned cloth tracking,'' \emph{arXiv preprint arXiv:2302.09502}, 2023.

\bibitem{longhini2025cloth}
A.~Longhini, M.~B{\"u}sching, B.~P. Duisterhof, J.~Lundell, J.~Ichnowski, M.~Bj{\"o}rkman, and D.~Kragic, ``Cloth-splatting: 3d cloth state estimation from rgb supervision,'' \emph{arXiv preprint arXiv:2501.01715}, 2025.

\bibitem{wang2024trtm}
W.~Wang, G.~Li, M.~Zamora, and S.~Coros, ``Trtm: Template-based reconstruction and target-oriented manipulation of crumpled cloths,'' in \emph{Proc. IEEE Int. Conf. Robot. Autom.}\hskip 1em plus 0.5em minus 0.4em\relax IEEE, 2024, pp. 12\,522--12\,528.

\bibitem{Edmund2025}
E.~Lo, X.~Huang, K.~Tang, A.~Seino, F.~Tokuda, and K.~Kosuge, ``Fabric flattening and alignment system using real-time mesh-based state estimation and visual servoing,'' in \emph{Proc. IEEE Int. Conf. Mechatron. Autom.}, 2025, pp. 328--334.

\bibitem{tang2024time}
K.~Tang, F.~Tokuda, A.~Seino, A.~Kobayashi, N.~C. Tien, and K.~Kosuge, ``Time-scaling modeling and control of robotic sewing system,'' \emph{IEEE/ASME Trans. Mechatron.}, vol.~29, no.~4, pp. 3166--3174, 2024.

\bibitem{yan2020learning}
W.~Yan, A.~Vangipuram, P.~Abbeel, and L.~Pinto, ``Learning predictive representations for deformable objects using contrastive estimation,'' \emph{CoRR}, vol. abs/2003.05436, 2020.

\bibitem{zhao2023learning}
T.~Z. Zhao, V.~Kumar, S.~Levine, and C.~Finn, ``Learning fine-grained bimanual manipulation with low-cost hardware,'' \emph{arXiv preprint arXiv:2304.13705}, 2023.

\bibitem{zhao2024aloha}
T.~Z. Zhao, J.~Tompson, D.~Driess, P.~Florence, K.~Ghasemipour, C.~Finn, and A.~Wahid, ``Aloha unleashed: A simple recipe for robot dexterity,'' \emph{arXiv preprint arXiv:2410.13126}, 2024.

\bibitem{black2024pi_0}
K.~Black, N.~Brown, D.~Driess, A.~Esmail, M.~Equi, C.~Finn, N.~Fusai, L.~Groom, K.~Hausman, B.~Ichter \emph{et~al.}, ``$\pi_0$: A vision-language-action flow model for general robot control,'' \emph{arXiv preprint arXiv:2410.24164}, 2024.

\bibitem{Ze2024DP3}
Y.~Ze, G.~Zhang, K.~Zhang, C.~Hu, M.~Wang, and H.~Xu, ``3d diffusion policy: Generalizable visuomotor policy learning via simple 3d representations,'' in \emph{Proc. Robotics Sci. Syst.}, 2024.

\bibitem{blender2025}
{Blender Online Community}, ``Blender -- a 3d modelling and rendering package,'' \url{https://www.blender.org}, Amsterdam, 2025, [Online; accessed September 4, 2025].

\bibitem{ravi2020accelerating}
N.~Ravi, J.~Reizenstein, D.~Novotny, T.~Gordon, W.-Y. Lo, J.~Johnson, and G.~Gkioxari, ``Accelerating 3d deep learning with pytorch3d,'' \emph{arXiv preprint arXiv:2007.08501}, 2020.

\bibitem{ravi2024sam2}
N.~Ravi, V.~Gabeur, Y.-T. Hu, R.~Hu, C.~Ryali, T.~Ma, H.~Khedr, R.~R{\"a}dle, C.~Rolland, L.~Gustafson, E.~Mintun, J.~Pan, K.~V. Alwala, N.~Carion, C.-Y. Wu, R.~Girshick, P.~Doll{\'a}r, and C.~Feichtenhofer, ``Sam 2: Segment anything in images and videos,'' \emph{arXiv preprint arXiv:2408.00714}, 2024.

\bibitem{Meshlab}
P.~Cignoni, M.~Callieri, M.~Corsini, M.~Dellepiane, F.~Ganovelli, and G.~Ranzuglia, ``{MeshLab: an Open-Source Mesh Processing Tool},'' in \emph{Eurographics Italian Chapter Conference}, V.~Scarano, R.~D. Chiara, and U.~Erra, Eds.\hskip 1em plus 0.5em minus 0.4em\relax The Eurographics Association, 2008.

\bibitem{cadene2024lerobot}
R.~Cadene, S.~Alibert, A.~Soare, Q.~Gallouedec, A.~Zouitine, S.~Palma, P.~Kooijmans, M.~Aractingi, M.~Shukor, D.~Aubakirova \emph{et~al.}, ``Lerobot: State-of-the-art machine learning for real-world robotics in pytorch,'' \url{https://github.com/huggingface/lerobot}, 2024.

\end{thebibliography}

\end{document}